\documentclass{article}
\usepackage[final]{pdfpages}
\usepackage{color}

\usepackage{spconf,amsmath,graphicx}
\usepackage{float}
\usepackage{subfig}
\usepackage{comment} 

\title{Two view constraints on the epipoles from few  correspondences}
\twoauthors
  {Yoni Kasten}
	{
Weizmann Institute of Science, Israel\\
ykasten@weizmann.ac.il}
  {Michael Werman}
	{
The Hebrew University of Jerusalem, Israel\\
michael.werman@mail.huji.ac.il}
\begin{document}
\onecolumn
\begin{center}

\includegraphics[width=0.9\textwidth, angle=0]{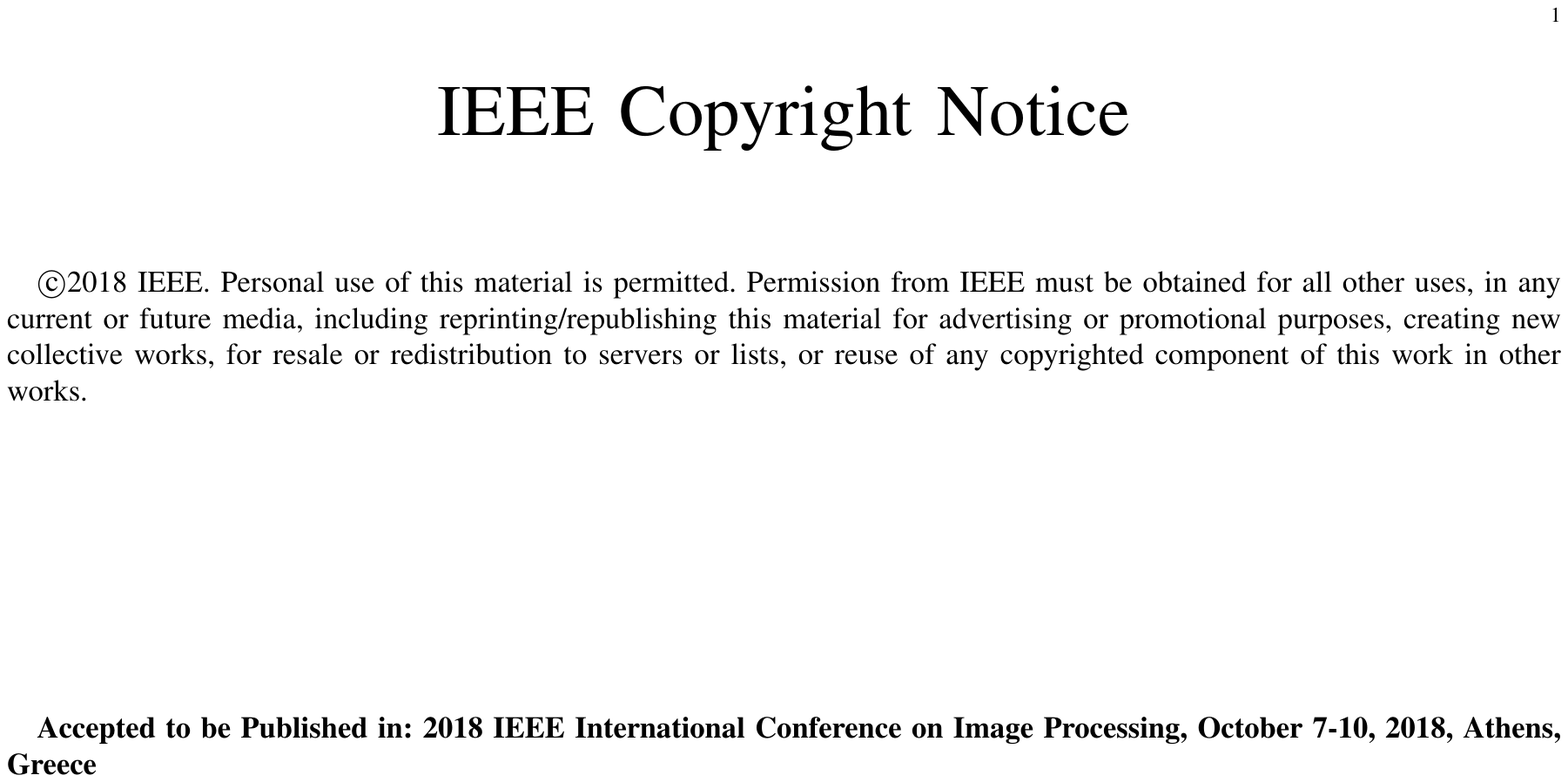}
\end{center}
\twocolumn
%
\maketitle
\begin{abstract}
In general it requires at least 7 point correspondences to compute the fundamental matrix between  views. We use the cross ratio invariance between corresponding epipolar lines, stemming from epipolar line homography, to derive a simple 
formulation for the  relationship between  epipoles and  corresponding points. 
We show how it can be used to reduce the number of required points for the epipolar geometry when some information about the epipoles is available and  demonstrate this with a buddy search app. 

\end{abstract}
\makeatletter
\def\blfootnote{\xdef\@thefnmark{}\@footnotetext}
\makeatother
\blfootnote{This research was supported by the Israel Science Foundation and by the  Israel Ministry of Science and Technology.}
\section{Introduction}
The fundamental matrix is the basic building block of multiple view geometry and its computation is the first step in many vision tasks. It is  usually computed from pairs of corresponding points. 
The best-known algorithm, for the fundamental matrix, is the eight points algorithm by Longuet-Higgins \cite{longuet1981computer}.  The {\it eight} point correspondences can be relaxed to seven. This results in  a cubic  equation with  one or three real solutions.

The fundamental matrix can also be computed from three matching epipolar lines \cite{hartley2003multiple} or equivalently the epipoles and three corresponding pairs of points. Given three such epipolar line correspondences, the one dimensional homography between the lines can be recovered. The 3 degrees of freedom for the 1-D homography together with the 4 degrees of freedom of the epipoles yield the required 7 degrees of freedom needed to compute the fundamental matrix. 
A few papers directly search for corresponding epipolar lines to compute the epipolar geometry,  \cite{sinha2010camera,ben2016epipolar,kasten2016fundamental,Calibration2016CVPR}.
\begin{figure}	
\centering
\includegraphics[width=0.7\linewidth]{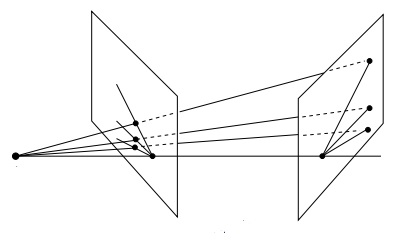}
\caption{The projection from any point on the baseline,  gives rise to the homography, $F[e]_\times$, between the epipolar line pencils}
\label{fig:1d-hom}
\end{figure}

In typical two view scenarios, points correspondences can be 
found.
But, as the angle between the views gets bigger it is more difficult to find corresponding points by automatic methods. The extreme case is when two cameras  face each other. In this case, correspondences can be found manually or by using known landmarks, 
and requiring fewer correspondences can be crucial. 

\begin{figure*}[t!]
\subfloat[][]{\includegraphics[width=0.25\linewidth]{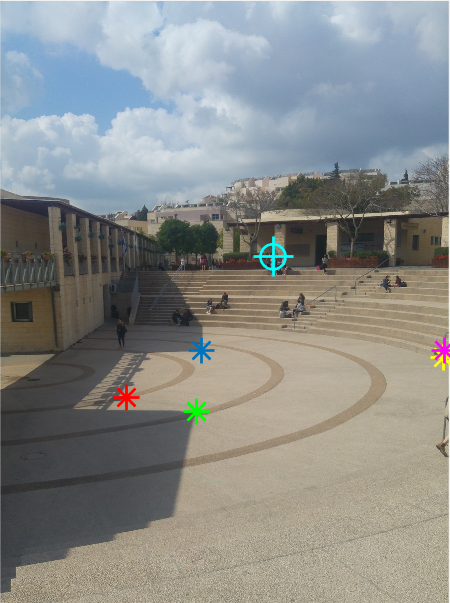}}
\subfloat[][]{\includegraphics[width=0.25\linewidth]{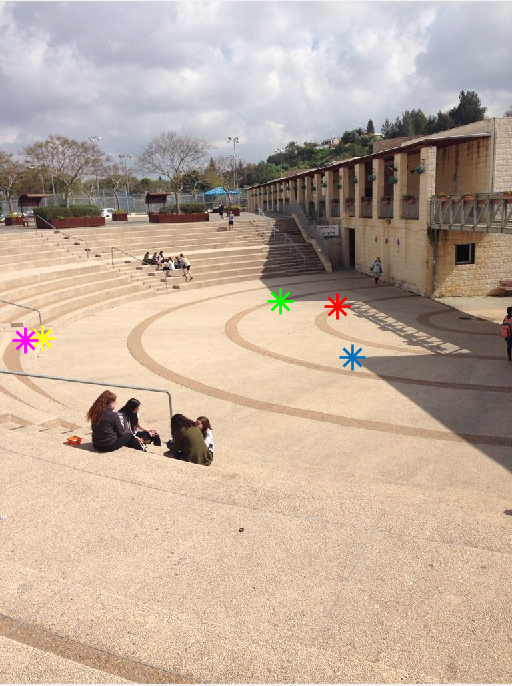}} 
\subfloat[][]{\includegraphics[width=0.25\linewidth]{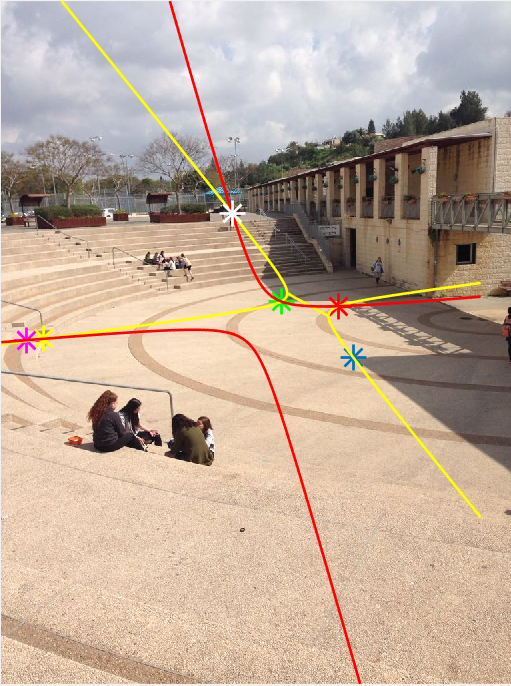}}
\subfloat[][]{\includegraphics[width=0.120\linewidth]{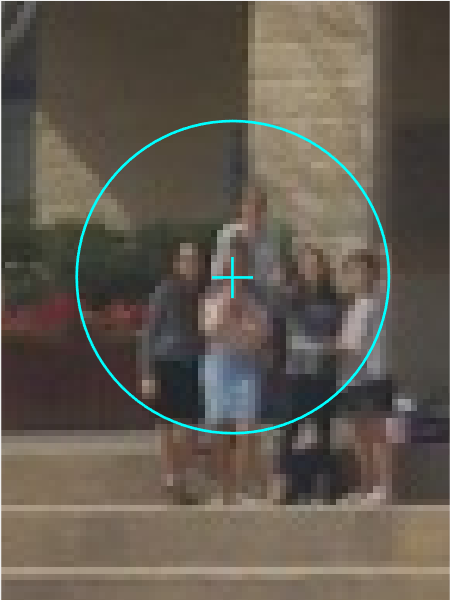}}   
\subfloat[][]{\includegraphics[width=0.120\linewidth]{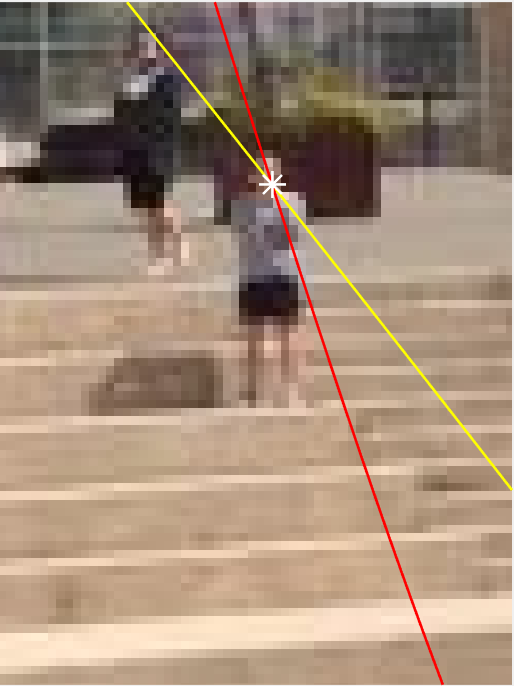}} 
\caption{
The typical setup where two cameras are facing each other. (a) and (b) are two images that are taken from two sides of a theater. Suppose that person (a) sees his friend (b) in the image, it means that one epipole is given. Five correspondences are marked by the friends, which we use to retrieve the epipolar geometry.  
(a) Image 1: 5 points of interest and the epipole are given. 
(b) Image 2: 5 corresponding points (marked with the same colors).
(c) Result: the epipole (white star) in Image 2 as calculated using our method. The red and the yellow curves are two conics resulted from 4 correspondences which passes through them and through the epipole. The intersection of the two conics is the epipole. 
(d), (e) Close ups on the input epipole in Image 1, and on the output epipole in Image 2 respectively.
}
\label{fig:buddy3}
\end{figure*}

When the cameras are facing each other, it may be possible to find  the other camera's position in the image, which is actually the epipole. Figure ~\ref{fig:buddy3} shows an example of a group of friends from two sides of a theater. If in one image, the epipole can be located, our method needs only 5 point correspondences to locate the other epipole and  the complete epipolar geometry,   

This is the motivation for using the epipole when it is available. In this paper we show how knowing one of the epipoles or even an epipolar line can be used to compute the epipolar geometry with less than 7 point correspondences.  

A series of papers \cite{chum2004epipolar,werner2001oriented,werner2003combinatorial} 
design  constraints stemming from the scene being in front of the camera, the existence of the plane at infinity and handedness of the camera system in the framework of oriented projective geometry and their related matroids. These constraints are then used to test the realizability of  configurations of putative corresponding point sets, where
\cite{nister2004four} treats the case of calibrated cameras.
\cite{Werner-cvpr03} uses the same setup to filter putative matches of 5 correspondences.
\cite{history} gives a historical perspective and pointers to many of the classic  results, where many of them were  originally introduced in \cite{german_1908}.

The main observation that this paper is based upon, is the existence of a 1-D homography between the corresponding pencils of  epipolar lines, see Figure~\ref{fig:1d-hom}. 
Each pencil being viewed as $\mathcal{P}^1$, 1-D projective space.
Every 4 corresponding epipolar lines, thus, have the same cross ratio. We develop the equations that follows from the cross ratio of the epipolar lines to develop a simple equation that we use in different configurations with less than 7 corresponding points.

\section{Corresponding points-epipole relation}

In this section we derive a simple relationship between epipoles  and corresponding points, based on matching epipolar lines.
 
Let $p_s\Leftrightarrow  p'_s$ be corresponding points and, $e$ and  $e'$  the epipoles  of a pair of images $I,I'$ then the cross-ratios of 4 corresponding epipolar lines 
$l_s=p_s \wedge e \Leftrightarrow l_s'=p'_s \wedge e'$
are equal:
\begin{align}
\frac{{\begin{vmatrix}l_1 l_2\end{vmatrix}}
{\begin{vmatrix}l_3 l_4\end{vmatrix}}}{{\begin{vmatrix}l_1l_3\end{vmatrix}}
{\begin{vmatrix}l_2 l_4\end{vmatrix}}}=
\frac{{\begin{vmatrix}l_1' l_2'\end{vmatrix}}
{\begin{vmatrix}l_3'l_4'\end{vmatrix}}}{{\begin{vmatrix}l_1'l_3'\end{vmatrix}}
{\begin{vmatrix}l_2' l_4'\end{vmatrix}}}
\end{align}
where 
$
{\begin{vmatrix}ab\end{vmatrix}}=\det{\begin{pmatrix}a_x&b_x\\
a_y&b_y\end{pmatrix}}
$.
Assigning $l_s=p_s \wedge e $ and $l_s'=p'_s \wedge e'$:
\begin{align} 
\label{equation:crossRatioEq}
\frac{{\begin{vmatrix}ep_1p_2\end{vmatrix}}
{\begin{vmatrix}ep_3p_4\end{vmatrix}}}{{\begin{vmatrix}ep_1p_3\end{vmatrix}}
{\begin{vmatrix}ep_2p_4\end{vmatrix}}}=
\frac{{\begin{vmatrix}e'p'_1p'_2\end{vmatrix}}
{\begin{vmatrix}e'p'_3p'_4\end{vmatrix}}}{{\begin{vmatrix}e'p'_1p'_3\end{vmatrix}}
{\begin{vmatrix}e'p'_2p'_4\end{vmatrix}}}
\end{align}
where 
$
{\begin{vmatrix}abc\end{vmatrix}}=\det{\begin{pmatrix}a_x&b_x&c_x\\
a_y&b_y&c_y\\
a_z&b_z&c_z\end{pmatrix}}
$.
Clearing denominators results in a conic in  the homogeneous elements of $e'$: $e'_x$, $e'_y$, and $e'_z$.

From this, it follows that given 4 point correspondences the epipoles are restricted to a 3-D manifold in $\mathcal{P}^4$ and if in addition one of the epipoles is given (2 more equations) the other epipole is restricted to lie on a conic (1-D manifold) of possible epipoles, $7-4-2=1$. With 5 point correspondences the epipoles are restricted to a 2-D manifold in $\mathcal{P}^4$ and if one of the epipoles is given (2 more equations) the other epipole is defined by a polynomial, 5'th degree Cremona\footnote{A polynomial transformation between projective spaces  which is defined everywhere and is bijective except perhaps
for points lying on a finite set of curves is called a Cremona transformation.} map, \cite{history}.
With 6 point correspondences, the pair of epipoles are restricted to a 1-D manifold in $\mathcal{P}^4$,
 the other epipole is directly computable by a polynomial, third degree Cremona map 
 \cite{history}.
 \vspace{-7mm}

\begin{figure*}[t!]
\centering

\subfloat[][]{
\includegraphics[width=0.3\linewidth]{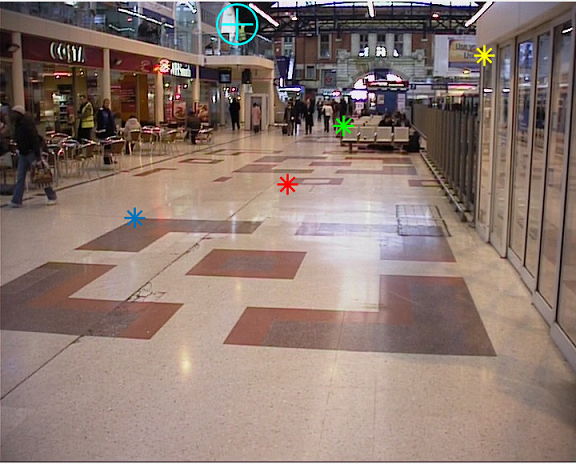}}
\subfloat[][]{
\includegraphics[width=0.3\linewidth]{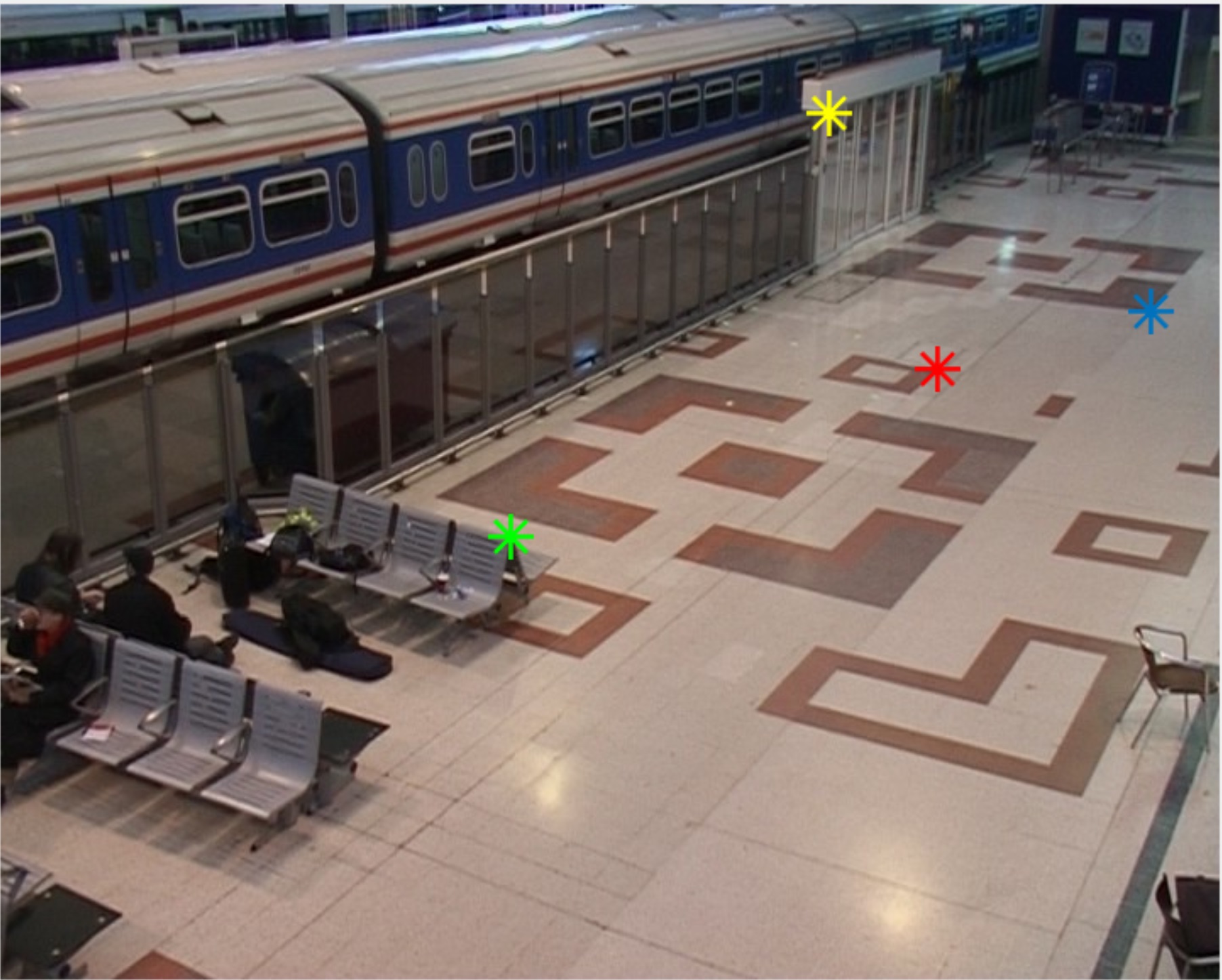}}\subfloat[][]{
\includegraphics[width=0.3\linewidth]{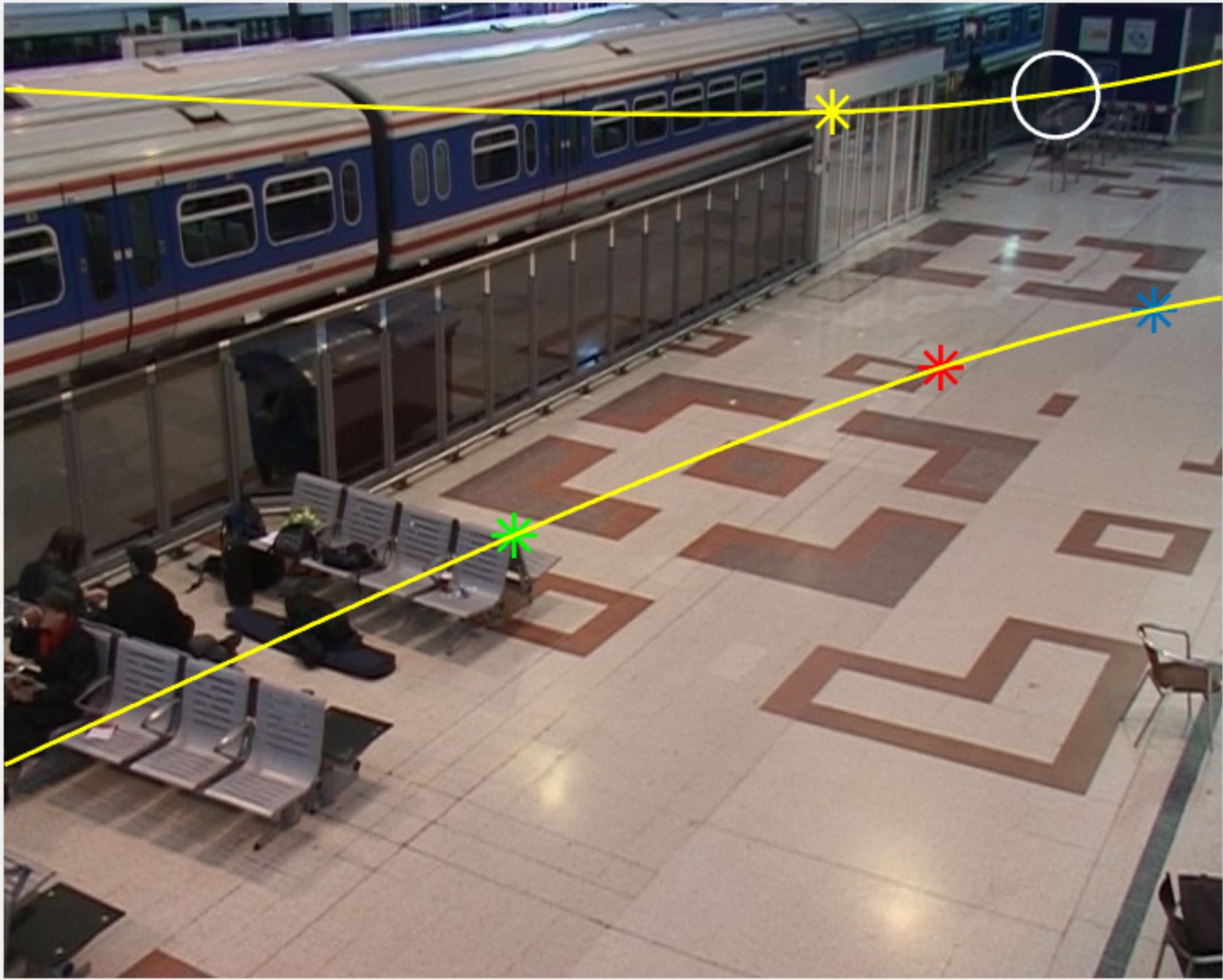}}

\caption{
Locating $e'$ given $e$ and 4 correspondences. 
(a) Image 1: 4 points of interest and $e$. 
(b) Image 2: 4 corresponding points (marked with the same colors) 
(c) The conic  for $e'$ is  in yellow (the 2 yellow curves are parts of the same conic). The  epipole is  in the white circle.
}
\label{fig:pets2006_4}
\end{figure*}

\vspace*{-1mm}
\section{Epipole localization with $n\leq 6$ correspondences}\label{section:epipoles_loc}
In this section we show how Equation~\ref{equation:crossRatioEq} can be used to calculate the other epipole given one epipole (or epipolar line) and less than 7 correspondences. Then, when the two epipoles are known, the epipolar geometry and the fundamental matrix between the views is completely known by using the epipolar lines homography. 
\subsection{4 Corresponding Points}
\label{section:4_points}
Given 4 points correspondences  between the images  and the epipole in Image 1,  Equation~\ref{equation:crossRatioEq} gives a single equation with two unknowns: the epipole's homogeneous  coordinates in Image 2: $e'$.  From Equation ~\ref{equation:crossRatioEq} we get:
\begin{align}
   \label{equation:crossRatio1Constraint}
{\begin{vmatrix}ep_ip_j\end{vmatrix}}
{\begin{vmatrix}ep_kp_l\end{vmatrix}}
{\begin{vmatrix}e'p'_ip'_k\end{vmatrix}}
{\begin{vmatrix}e'p'_jp'_l\end{vmatrix}}=
\nonumber
\\
{\begin{vmatrix}e'p'_ip'_j\end{vmatrix}}
{\begin{vmatrix}e'p'_kp'_l\end{vmatrix}}
{\begin{vmatrix}ep_ip_k\end{vmatrix}}
{\begin{vmatrix}ep_jp_l\end{vmatrix}}
\end{align}
Let  $e'=(e_x',e_y',e_z')^T$:
\begin{align}
ae_x'^2+be_x'e_y'+ce_y'^2+de_x'e_z'+ee_y'e_z'+fe_z'^2=0
\end{align}
Where $a,b,c,d,e,f$ are numerical coefficients that depend on the inputs.
This can be written as:
\begin{align}
(e_x',e_y',e_z')^TC(e_x',e_y',e_z')=0
\end{align}
Where $C=\left( \begin{array}{ccc}
 a & \frac{b}{2} & \frac{d}{2}\\
 \frac{b}{2}& c &  \frac{e}{2} \\
 \frac{d}{2} & \frac{e}{2} & f
\end{array} \right)$. As a result we got the conic $C$ in Image 2 on which $e'$ is located\footnote{The conic also passes through the points of interest in Image 2 since replacing $e'$ with $p_i'$ for example in Equation ~\ref{equation:crossRatio1Constraint}, zeros the determinants: ${\begin{vmatrix}e'p'_ip'_k\end{vmatrix}}$ and ${\begin{vmatrix}e'p'_ip'_j\end{vmatrix}}$}.

\subsection{5 Corresponding Points}\label{section:fivePoints}
When using 5 corresponding points, two equations of  form  Equation~\ref{equation:crossRatio1Constraint} can be used with  different subsets of 4 corresponding points. Given the epipole in Image 1: $e=(e_x,e_y,e_z)^T$, each equation of the form Equation~\ref{equation:crossRatio1Constraint} gives a conic in Image 2 that $e'$ and the 4 given points are incident to. The two conics intersect in 4 points: $e'$ and the 3 points which are the intersection of the two subsets. Note that the coordinates of these 3 points are already known.

There is a 5'th degree polynomial relating the epipoles
which in \cite{history, Werner-cvpr03} is attributed to Strum (1869).
The intersection of two conics sharing three points, which for simplicity
of the equation are chosen as $(1,0,0)^T$,  $(0,1,0)^T$ and  $(0,0,1)^T$ \footnote{The 3 points can can always be transformed thusly  with a 2D homography $H$ and at the end transformed back with $H^{-1}$.},  can be computed using the reciprocal Cremona transformation. The reciprocal to $(x,y,z)$, $(x,y,z)^*$ is 
$ (yz,zx,xy)$, (projectively the same as 
$(\frac{1}{x},\frac{1}{y},\frac{1}{z})$).
 The fourth common point of two conics each presented by 2 more points on the conics:
$\{x_1, x_2\}$ and $\{y_1, y_2\}$ is 
$[(x_1^* \vee x_2^*) \wedge ( y_1^* \vee y_2^*)]^*$,
where $\vee$ is the line connecting two points and $\wedge$ is the point of intersection of two lines. 

\begin{figure*}
\centering

\subfloat[][]{
\includegraphics[width=0.3\linewidth]{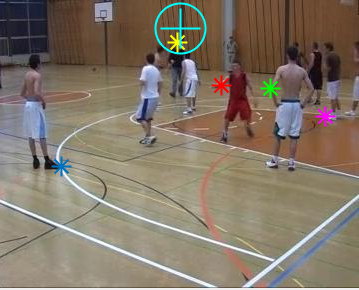}}
\subfloat[][]{
\includegraphics[width=0.3\linewidth]{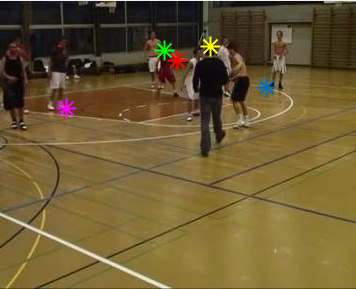}}\subfloat[][]{
\includegraphics[width=0.3\linewidth]{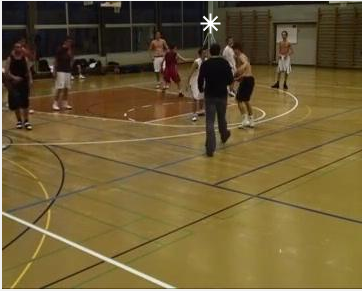}}

\caption{
Finding $e'$ given $e$ and 5 correspondences,  tested on EPFL basketball dataset, Cameras 1 and 3.
}
\label{fig:epfl}
\end{figure*}
\subsection{6 Corresponding Points}
\label{section:sixPoints}
When using 6 corresponding points between two views, 3 independent equations of the form of Equation~\ref{equation:crossRatio1Constraint} can be constructed from three different subsets of 4 corresponding points, so the epipoles are on a 1-D curve in $\mathcal{P}^4$. Given a 1-D curve that the epipole is  on in one of the images (for example epipolar line in Image 1), the epipoles of both views can be computed. 

There is a 3'th degree polynomial relating the epipoles
which in \cite{history} is attributed to Strum (1869).

\section{Applications}

\label{section:buddy}
Finding someone in a crowd is often a daunting task.
We consider a scenario of 2 buddies searching for each other at the same venue as is common in concerts and sport games. Other cases are military scenarios where it is crucial for a fighter to locate his partner in the scene.   

In these cases the angle between the cameras can be large and automatically finding corresponding feature points is hard. Using the constraints that are defined in Section~\ref{section:epipoles_loc} makes the mission simpler since it requires less corresponding  points. We demonstrate such scenarios in Section~\ref{section:buddy_expr}. We implemented it in a cellular app which will be made publicly available. 
\vspace*{-3mm}
\section{Experiments}

\subsection{Epipole localization}
As described in section~\ref{section:epipoles_loc},   $n<7$ corresponding points between the views
can be used to help in searching for the epipoles.
We tested our method on several public datasets.


\subsubsection{4 Corresponding Points}
In this setup, 4 corresponding points between the images are given, and the epipole in Image 1 is given as well. The method from Section ~\ref{section:4_points} was used for locating the second epipole.

Figure~\ref{fig:pets2006_4} shows the conic computed using this method on images from Cameras 1 and 4 in the Pets 2006 dataset which face each other. The true epipole in Image 2 is in the top-right corner of the image. The conic  passes through the epipole. 

\subsubsection{5 Corresponding Points}
In this setup, 5 corresponding points between the images are given, and the epipole in Image 1 is given as well. The method from Section \ref{section:fivePoints} was used for locating the second epipole.

Figure~\ref{fig:epfl} show an example on a frame from EPFL Basketball dataset \cite{Berclaz11}. given 5 corresponding points and the epipole  in Image 1, the epipole  in Image 2 is successfully calculated.

\subsubsection{6 Corresponding Points}



In this setup, 6 corresponding points are given. The method from Section \ref{section:sixPoints} was used for locating the two epipoles.

\begin{figure}[t!]
\centering
\includegraphics[width=0.48\linewidth]{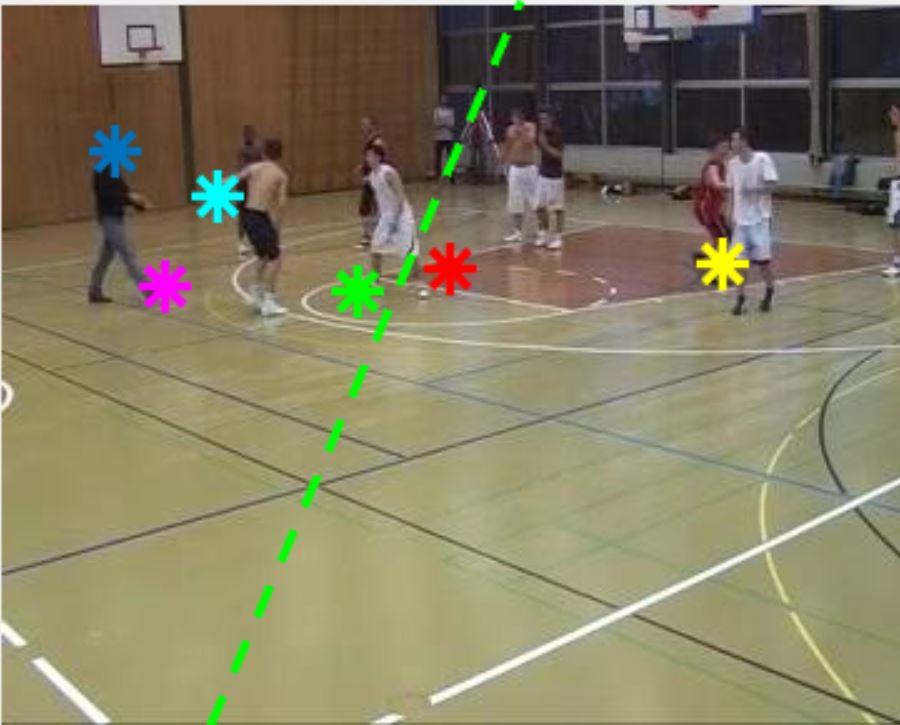} 
\includegraphics[width=0.48\linewidth]{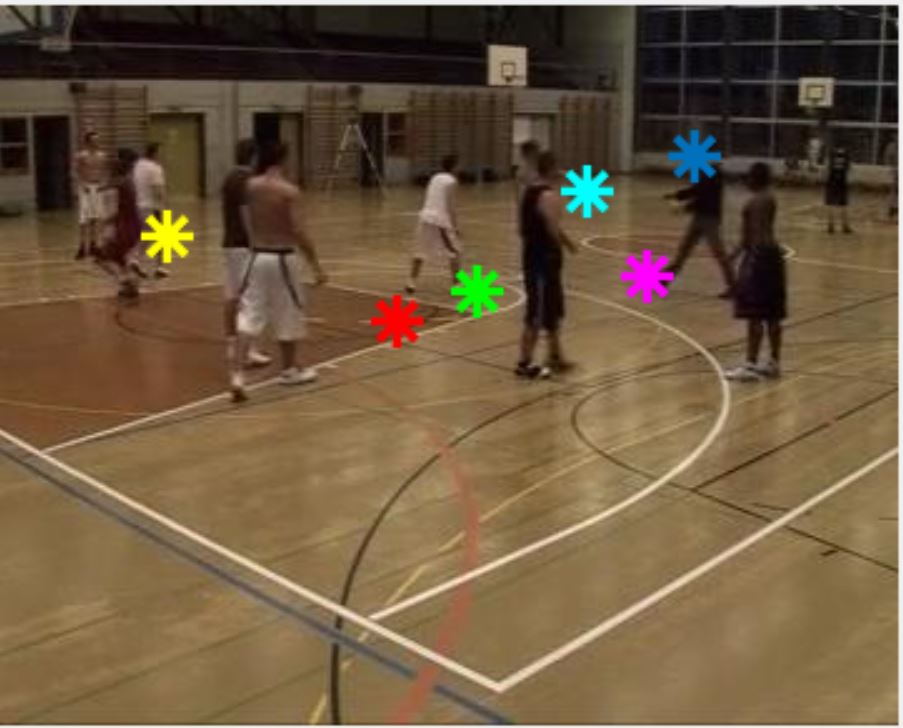} \\
(a)\\
\includegraphics[width=0.48\linewidth]{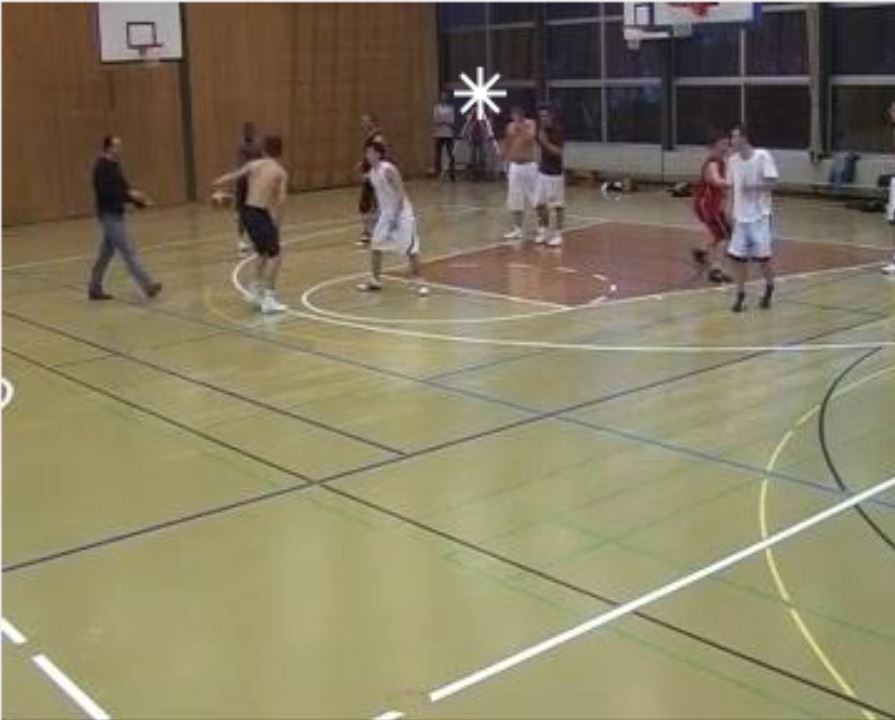}  
\includegraphics[width=0.48\linewidth]{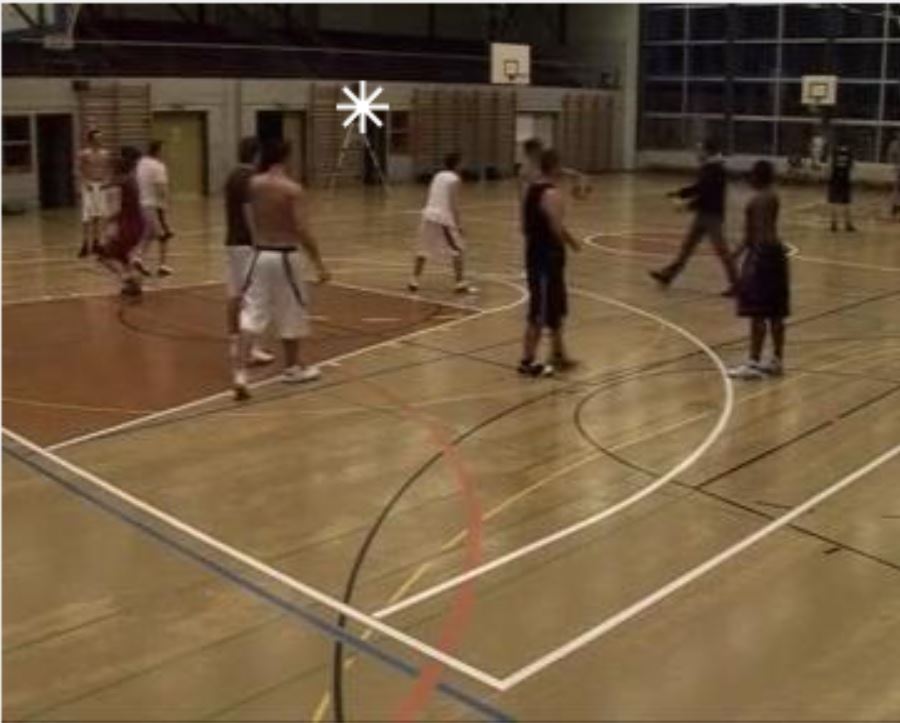} 

(b)

\caption{
Locating $e'$ and $e$ given 6 correspondences and a epipolar line in Image 1. (EPFL basketball dataset, Cameras 2 and 4). 
(a) The inputs: in Image 1 (left) 6 points of interest and an epipolar line (dashed green line) that is  passing over the leg of the second camera. In Image 2 (right): 6 corresponding points marked with the same colors. 
(b) $e$ and $e'$ are successfully located  and marked white.
}
\label{fig:sixPoints}
\end{figure}

Given epipolar line in Image 1 (epipole location up to 1 degree of freedom), the two epipoles can be computed accurately.
Figure ~\ref{fig:sixPoints} shows the results for Cameras 2 and 4 in the EPFL Basketball dataset.

\subsection{Buddy Search}
\label{section:buddy_expr}
\vspace{-0.3cm}
As described in Section~\ref{section:buddy} a practical application of our method is buddy finding. 
An example of a group of friends, which locate their friend in the other side of the theater by using the application, is given in Figure~\ref{fig:buddy3}.
\vspace*{-3mm}
\section{Conclusions}

We presented a method to exploit knowledge on the location of the epipole given only 4,5 or 6 points correspondences. This results in constraints on the epipolar geometry and can simplify the search for the epipolar points or even to recover  the complete epipolar geometry. 

 \clearpage
 \newpage
\bibliographystyle{IEEEbib}
\bibliography{sample}

\end{document}